\documentclass{arxiv}
\pdfoutput=1

\usepackage{amsmath,amsfonts,amssymb}
\usepackage{changes}
\usepackage{url}
\usepackage{graphicx}
\usepackage{times}

\newcommand{\rev}[1]{#1} 

\title{Adaptive and Resilient Soft Tensegrity Robots}
\date{}
\author{John Rieffel$^{1,*}$ and Jean-Baptiste Mouret$^{2,3,4,\ast,\dagger}$\\
~\\
\normalsize{$^1$ Union College, Schenectady, NY 12308, USA}\\
\normalsize{$^2$ Inria Nancy Grand - Est, Villers-l\`es-Nancy, F-54600, France}\\
\normalsize{$^3$ CNRS, Loria, UMR 7503,  Vand\oe{}uvre-l\`es-Nancy, F-54500, France}\\
\normalsize{$^4$ Universit\'e  de Lorraine, Loria, UMR 7503, Vand\oe{}uvre-lès-Nancy, F-54500, France}\\
~\\
\normalsize{$^\ast$ J.R. and J.-B. M. contributed equally to this work}\\
~\\
\normalsize{$\dagger$To whom correspondence should be addressed; E-mail:{\texttt{rieffelj{@}union.edu}}}
}

\begin{document}
\frenchspacing
\maketitle

\normalfont
\sffamily\bfseries
\noindent{}Living organisms intertwine soft (e.g., muscle) and hard (e.g., bones) materials, giving them an intrinsic flexibility and resiliency often lacking in conventional rigid robots. The emerging field of soft robotics seeks to harness these same properties in order to create resilient machines. The nature of soft materials, however, presents considerable challenges to aspects of design, construction, and control -- and up until now, the vast majority of gaits for soft robots have been hand-designed through empirical trial-and-error.  This manuscript describes an easy-to-assemble tensegrity-based soft robot capable of highly dynamic locomotive gaits and demonstrating structural and behavioral resilience in the face of physical damage.  Enabling this is the use of a machine learning algorithm able to discover effective gaits with a minimal number of physical trials.  These results lend further credence to soft-robotic approaches that seek to harness the interaction of complex material dynamics in order to generate a wealth of dynamical behaviors.

\normalfont

\section*{Introduction}

Unlike machines, animals exhibit a tremendous amount of resilience, due in part to their intertwining of soft tissues and rigid skeletons.  In nature, this suppleness leads to several compelling behaviors which exploit the dynamics of soft systems.  Octopi, for example, are able to adaptively shape their limbs with ``joints'' in order to
perform efficient grasping~\citep{sumbre2005neurobiology}. Jellyfish
exploit their inherent elasticity in order to passively recover energy
during swimming~\citep{gemmell2013passive}.  {\em Manduca sexta}
caterpillars have a mid-gut which acts like a ``visceral-locomotory
piston'' -- sliding forward ahead of the surrounding soft tissues,
shifting the animal's center of mass forward well before any visible
exterior change~\citep{simon2010visceral}.

\begin{figure}
\centering\includegraphics[width=\linewidth]{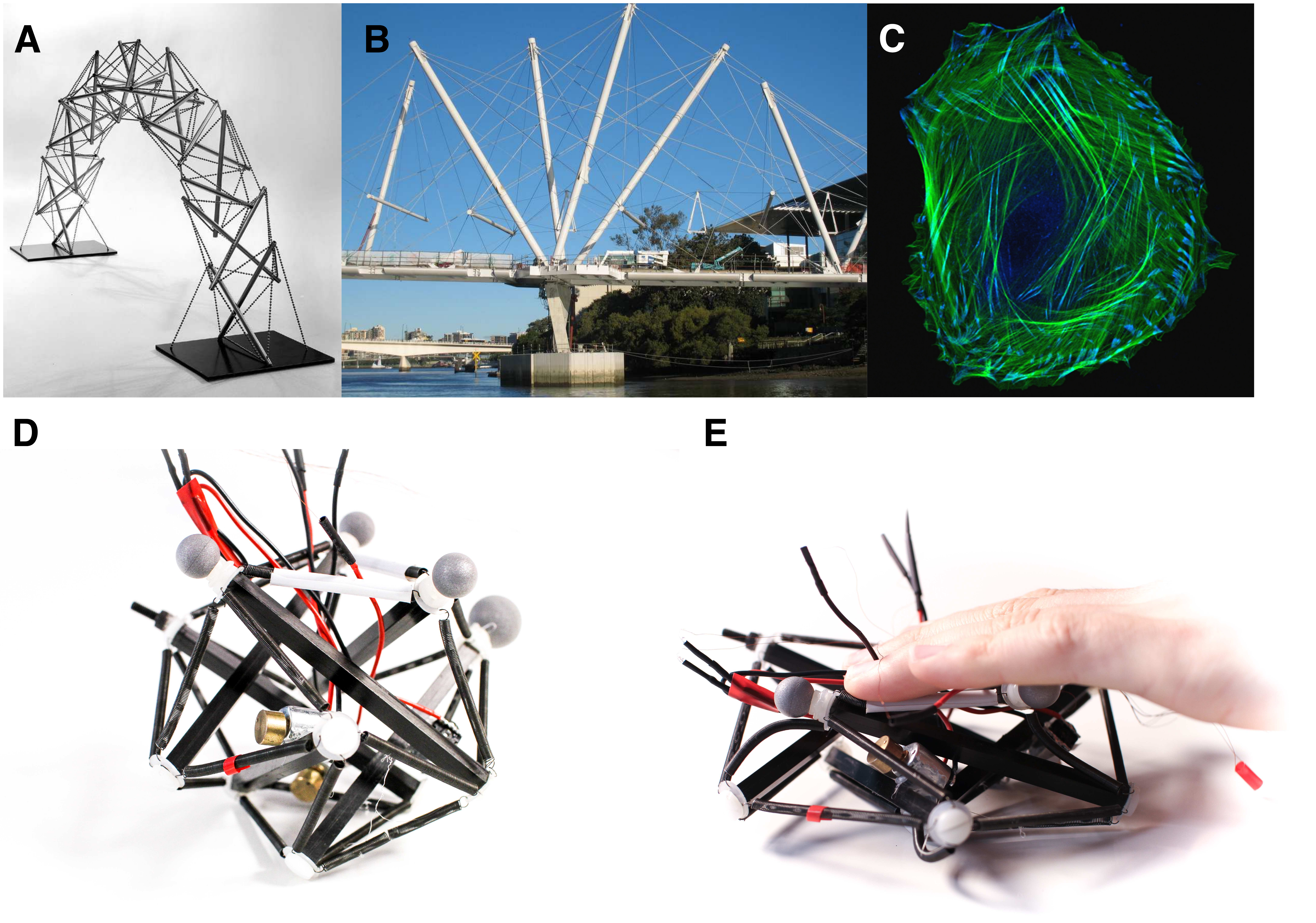}
\caption{\label{fig:concept}\textbf{Concept of our soft tensegrity robot.} Tensegrity structures are combinations of rigid elements (struts) joined at their endpoints by tensile elements (spring or cables) that are kept stable by the interplay of pre-stress forces. \textbf{A}. The first tensegrity structures appeared in art, with the sculptures of Kenneth Snelson~ \citep{snelson2012,skelton2009tensegrity}. \textbf{B.} They have been subsequently used in architecture, for instance for the Kurilpa bridge (Brisbane, Australia). \textbf{C}. More recently, tensegrity has been found to be a good model of the mechanotransduction of living cells~ \citep{wang2001mechanical}. \textbf{D.} Our tensegrity robot is based on carbon struts and springs. It is is actuated by 3 vibrators (glued to 3 of the struts) whose frequency is automatically tuned by a trial-and-error learning algorithm (Methods). \textbf{E.} Thanks to the tensegrity structure and to the compliance of the springs, our robot will keep its integrity when deformed and spring back into its initial form. A video is available in supplementary materials (Video S1 --- \protect\url{https://youtu.be/SuLQDhrk9tQ}).}
\end{figure}

Taking inspiration from the natural world, the field of soft robotics seeks to address some of the constraints of conventional rigid robots through the use of compliant, flexible, and elastic materials  \citep{lipson2014challenges,wehner2016integrated}. Trimmer {\em et al.}, for instance, construct soft robots from silicone
rubber, using  shape-memory alloy (SMA) micro-coil actuation, which
can slowly crawl in controlled fashion~ \citep{trimmer-ieee} or roll in an
uncontrolled ballistic fashion~ \citep{lin2011}.  Similarly, research by Whitesides {\em et al.} uses pneumatic inflation to produce slow,
dynamically stable crawling motions  \citep{shepherd2011multigait} as
well as fast, but less controlled tentacle-like grippers  \citep{martinez2013robotic}, combustion-driven jumpers  \citep{bartlett20153d} and a self-contained microfluidic ``octobot''  \citep{wehner2016integrated}.

Despite their advantages, soft-material robots are difficult to control by conventional means  \citep{lipson2014challenges,shepherd2011multigait}.
They are by their very nature high dimensional dynamic systems with an essentially infinite number of degrees of freedom.
The elasticity and deformability which provide their appeal come at the cost of resonances and tight dynamic coupling between components  \citep{trimmer-ieee}, properties which are often avoided, or at least suppressed, in conventional engineering approaches to robotic design.  This complexity precludes the use of many of the traditional kinematic and inverse-dynamics approaches to robotic control  \citep{craig2007}.

As a result, up until now, the locomotive gaits of most soft robots have been developed by hand through empirical trial-and-error  \citep{shepherd2011multigait}.
This process can be both challenging and time consuming, particularly when seeking to fully exploit the dynamical complexity of soft mechanisms.
Importantly, this manual process also prevents these robots from adapting their control strategy when the context changes, for instance when they encounter an unexpected type of terrain, or when they are physically damaged.





In this work, we describe a new class of soft robot based upon a tensegrity structure driven by vibration. Like many other soft robots, this tensegrity robot is resilient, and can resist damage when perturbed or crushed.  Unlike other soft robots, however, this particular modular tensegrity robot is easy to build, easy to control, and thanks to a data-efficient reinforcement learning algorithm  \citep{cully2015robots}, it can autonomously discover how to move, and  quickly relearn and adapt its behavior when damaged.

\rev{ Vibration is an increasingly popular method of sensor-free manipulation and control for automated systems~  \citep{berretty2001trap}.   Rezik {\em et al.}, for instance, developed a vibration-driven planar manipulator~  \citep{reznik1998coulomb} able to perform large-scale distributed planar control of small parts~ \citep{reznik2000building}.   In mobile robotics, stick-and-slip frictional motion \citep{vartholomeos2006analysis,joe2015vibration}  driven by paired vibrating motors has been used in a variety of mobile robots \citep{rubenstein2012kilobot,parshakova2016ratchair}.  Often, these approaches use empirically-derived hand-tuned frequencies to generate motion, using linear interpolation of their two motor speeds in order to smoothly generate a range of behaviors.  One weakness of vibration-based approaches to locomotion is that vibration of this type leads to unpredictable motion, even when assuming perfectly consistent surfaces \citep{joe2015vibration}, which presents a challenge to modeling and simulation.}


Tensegrities  are relatively simple mechanical systems, consisting of a number of rigid elements (struts) joined at their endpoints by tensile elements (cables or springs), and kept stable through a synergistic interplay of pre-stress forces (Fig.~\ref{fig:concept}~A-C). Beyond engineering, properties of tensegrity have been demonstrated at all scales of the natural world, ranging from the tendinous network of the human hand  \citep{valero2007tendon} to the mechanotransduction of living cells  \citep{wang2001mechanical}. At every size, tensegrity structures exhibit two interesting features  \citep{snelson2012,skelton2009tensegrity}: they have an impressive strength-to-weight ratio, and they are structurally robust and stable in the face of deformation.
Moreover, unlike many other soft robots, tensegrity structures are inherently modular (consisting of only struts and springs) and are therefore relatively easy to construct.
They are simple enough to be baby toys and featured in books for children activities  \citep{ceceri2015making}, while complex enough to serve as the basis for the next generation of NASA's planetary rovers  \citep{caluwaerts2014design}.

The most common control method for tensegrity robots is to slowly change the lengths of the struts and/or cables, causing large-scale, quasi-static (rather than dynamic) structural deformations, which, in turn, make the robot move via tumbling and rolling~  \citep{caluwaerts2014design, koizumi2012rolling}. As they assume that the structure is relatively stiff throughout locomotion, such control strategies are not suitable for more compliant soft tensegrity robots. In addition, they lead to slow locomotion speeds.

Lately, researchers have begun investigating more dynamical methods of tensegrity robot control.   Bliss {\em et al.} have used central pattern generators (CPGs) to produce resonance entrainment of simulated non-mobile  tensegrity structures~  \citep{bliss2012resonance}.  Mirletz {\em et al.} have used CPGs to produce goal-directed behavior in simulated tensegrity-spine-based  robots~  \citep{mirletz2015goal}.    These efforts, however valuable, were all produced in simulated environments, and have not yet been successfully transferred into real-world robots.  As Mirletz {\em et al} point out~  \citep{bliss2012resonance}, the dynamic behavior of tensegrities is highly dependent upon the substrate they interact with -- this means that results developed in simulated environments cannot necessarily be simply transferred to real robots (in Evolutionary Robotics, this is known as the ``Reality Gap''~  \citep{jakobi1995noise,koos2013transferability}).

\rev{More recently, B{\"o}hm and Zimmermann developed a tensegrity-inspired robot actuated by an single oscillating electromagnet~  \citep{bohm2013vibration}.  Although this robot was not a pure tensegrity (it rigidly connected multiple linear struts), it was, compellingly, able to change between forward and backward locomotion by changing the frequency of the oscillator.   Vibration has been proposed as a means of controlling much softer robots as well~  \citep{berger2015growing}.}

Here we explore the hypothesis that the inherent resonance and dynamical complexity of real-world soft tensegrity robots can be beneficially harnessed (rather than suppressed), and that, if properly excited  \citep{oppenheim2001vibration}, it can resonate so that the robot performs step-like patterns that enable it to locomote. To test this hypothesis and demonstrate the potential of soft tensegrity robots, we designed a pocked-size, soft tensegrity robot whose parameters were tuned to maximize resonance, and whose goal is to locomote as fast as possible across flat terrain. To find the right vibrational frequencies, we equipped the robot with a data-efficient trial-and-error algorithm, which also allows it to adapt when needed.

\section*{Results}







\begin{figure}
\centering
\includegraphics[width=\linewidth]{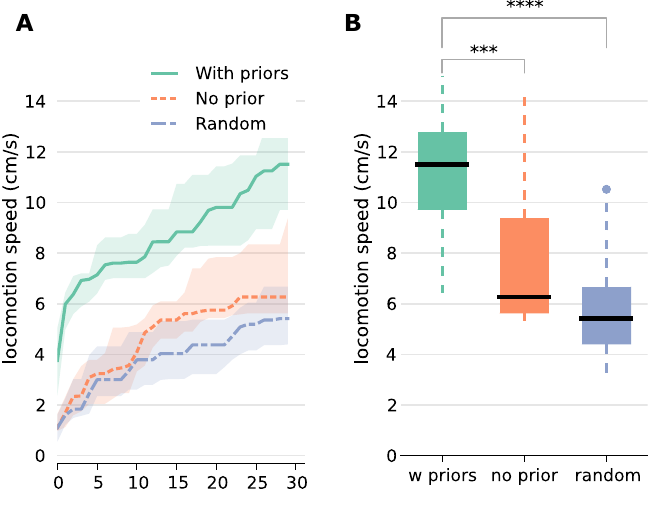}
\caption{\label{fig:learning-res}\textbf{Evaluation of the learning algorithm.} \textbf{A. Locomotion speed after each of the 30 trials}. The light zones represent the 25\textsuperscript{th} and 75\textsuperscript{th} percentiles. \textbf{B. Locomotion speed after 30 trials.} The central mark is the median, the edges of the box are the 25\textsuperscript{th} and 75\textsuperscript{th} percentiles (inter-quartile range -- IQR), the whiskers corresponds to the range $[25\% - 1.5 \times IQR, 75\% + 1.5\times IQR]$, and points outside of the whiskers are considered to be outliers (this corresponds to the ``interquartile rule''). Each condition is tested with 20 independent runs of the algorithms.}
\end{figure}

Our soft tensegrity robot (Fig.~\ref{fig:concept}D-E) is based upon a canonical six-bar tensegrity shape consisting of  equal length composite struts  connected  via 24 identical helical springs, with four springs emanating from each strut end.  Unlike most tensegrity structures, which seek to maximize stiffness \rev{by, among other things,  using nearly inelastic cables  \citep{oppenheim2001vibration}, here we replace the cables with very elastic springs, with spring constants chosen} with the goal of producing suitably low natural frequencies of the structure, with corresponding large displacements -- in other words, to maximize suppleness.  This allows the pocket-sized robot to maintain its structural shape under normal operation, and yet be easily compressed flat in one's hand.  A variable speed motor coupled to offset masses was then attached to three of the struts in order to excite the natural frequencies of the structure.  \rev{The motor and weight were chosen to be in a range consistent with preliminary models.  Because of the difficulty in modeling real-world interactions of these tensegrity robots, as well as the fact that we use a real-world optimization process described below, variability in the exact placement of each motor on a strut is allowed. }

Like many robots, the tensegrity robot needs to use different gaits in order to achieve lomocotion, depending on terrain. In our case, these gaits are determined by the speeds of the three vibratory motors.    As the exact properties of the terrain are seldom known {\em a priori}, and  because hand-designing gaits is time consuming (not to mention impossible when the robot is in remote or hazardous environments) this robot finds effective motor frequencies by using a trial-and-error learning algorithm whose goal is to maximize the locomotion speed.

Earlier work of ours~  \citep{khazanov2013exploiting, khazanov2014evolution} used interactive trial-and-error as well as automated hill climbing techniques to find optimal gaits for a tensegrity robot.  These gaits, could in turn, be incorporated into a simple state machine for directional control. However, these techniques required hundreds of physical trials that were  time consuming and produced significant wear on the physical robot.  More importantly, the interactive procedure required a human in the loop, whereas we envision robots that can adapt autonomously to new situations (e.g. a damage or a new terrain).  \rev{The work described in this paper, by automating the optimization process while minimizing the number of physical trials required, substantially reduces the amount of human interaction required, and is an important step toward full autonomy.}

Here, as a substantial improvement upon these earlier time-intensive methods, we employ a Bayesian optimization algorithm  \citep{cully2015robots,ghahramani2015probabilistic,shahriari2016taking}, which is a mathematical optimizer designed to find the maximum of a performance function with as few trials as possible.

Conceptually, Bayesian optimization fits a probabilistic model (in this case a Gaussian process  \citep{Rasmussen2006}, see Methods) that maps motor speeds to locomotion speed. Because the model is probabilistic, the algorithm can not only predict which motor speeds are the most likely to be good, but also associate it to a confidence level. Bayesian optimization exploits this model to select the next trial by balancing exploitation -- selecting motor speeds that are likely to make the robot move faster -- and exploration -- trying combinations of motor speeds that have not been tried so far (Methods). As an additional benefit, this algorithm can take into account that observations are by nature uncertain.

\begin{figure}
\centering
\includegraphics[width=0.9\linewidth]{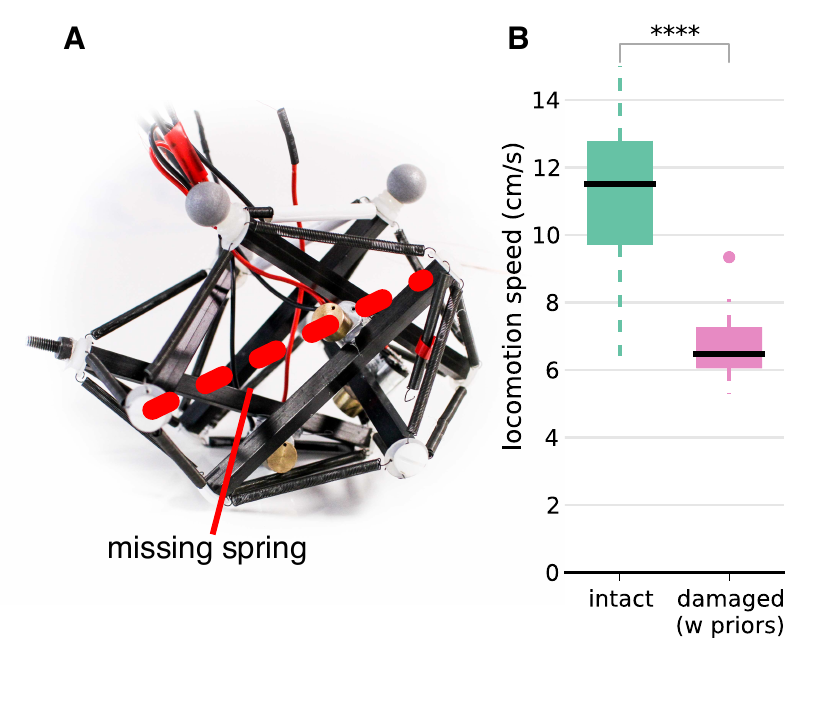}
\caption{\label{fig:broken-res}\textbf{Experiments with a damaged robot.} \textbf{A. Damaged robot.} A spring is disconnected from the robot of Fig.~\ref{fig:concept}. \textbf{B. Locomotion speed after 30 trials}. The central mark is the median, the edges of the box are the 25\textsuperscript{th} and 75\textsuperscript{th} percentiles (inter-quartile range -- IQR), the whiskers corresponds to the range $[25\% - 1.5 \times IQR, 75\% + 1.5\times IQR]$, and points outside of the whiskers are considered to be outliers (this corresponds to the ``interquartile rule''). Each condition is tested with 20 independent runs of the algorithms.}
\end{figure}

The Bayesian optimization algorithm usually starts with a constant prior for the expected observation (e.g., the expected speed is 10 $cm/s$) and a few randomly chosen trials to initialize the model.
For this robot, however, common sense, along with preliminary modeling, suggests that speeds near the motor maximums are more likely to produce successful gaits, and that near-zero motor speeds are not expected to make the robot move.  This insight was substantiated in preliminary experiments: many effective gaits were produced by high motor speeds, both forward and backward. Therefore, to speed up learning, we  use a  non-linear prior model as follows: (i) if the three motor speeds are close to 0, then we should expect a locomotion speed close to 0 and (ii) if all the motors are close to full speed (in any direction), then we should expect the maximum locomotion speed (Methods and Fig.~\ref{fig:profiles}D).  Thanks to this prior, the Bayesian optimization algorithm does not need any random sample points to seed the prior and instead starts with promising solutions. In spite of this prior learning is still needed, because many combinations of motors at full speeds make the robot tumble or rotate on itself, resulting in low performance; in addition, subtle changes to motor speeds can have dramatic effects upon the resulting robot gait. 

\begin{table*}
  \sffamily

\begin{center}
\begin{tabular}{ll}
 \hline
 Weight (W) & $89 g$\\
 Body length & $13 cm$ \\
 Best locomotion speed (S) [after learning]& $15 cm.s^{-1}$ ($1.15$ body lengths per second)\\
 Current drawn (full speed): & $700 mA$ \\
 Power drawn at 5V (P) & $3.5 W$\\
 Cost of transport at maximum speed (COT) & $262$ ($\textrm{COT} \triangleq \frac{P}{W S}$)\\
  \hline
\end{tabular}
\end{center}
\caption{\label{table:data}Best locomotion speed, power consumption, and  cost of transport for the gait that corresponds to the maximum speed at 5V. For reference, a cost of transport (COT) of 262 is comparable to the COT of a mouse (more than 100), but much higher than a car or a motorcycle (around 3) \citep{tucker1970energetic}.}
\end{table*}


We first evaluate the effectiveness of the learning algorithm (Fig.~ \ref{fig:learning-res}). The performance function is the locomotion speed, measured over 3 seconds, in any direction. If the robot turns too much, that is if the yaw exceeds a threshold, the evaluation is stopped (Methods). The covered distance is measured with an external motion capture system (Methods), although similar measurements can be obtained with an onboard visual odometry system  \citep{cully2015robots,davison2007monoslam}. We compare three algorithms: random search, Bayesian optimization without prior (using 10 random points to initialize the algorithm), and Bayesian optimization with prior. Each algorithm is allowed to test 30 different motor combinations (resulting in 90 seconds of learning for each experiment) and is independently run 20 times to gather statistics. The results show that the best locomotion speeds are obtained with the prior-based Bayesian optimization ($11.5 cm/s$, 5\textsuperscript{th} and 95\textsuperscript{th} percentiles $[8.1, 13.7]$), followed by the prior-free Bayesian optimization ($6.3 cm/s [5.5, 12.4]$). The worst results are obtained with the random search ($5.4 cm/s [3.5, 9.9]$). \rev{The absolute best locomotion speed ($15 cm.s^{-1}$ was found with Bayesian optimization (Table \ref{table:data}) and corresponds to $1.15$ body lengths per second.} Overall, these experiments demonstrate that the prior-based Bayesian optimization is an effective way to automatically discover a gait in only $30$ trials with this robot. Videos of a a typical gait is available as supplementary material (Video S1).

\begin{figure*}
\centering\includegraphics[width=0.7\linewidth]{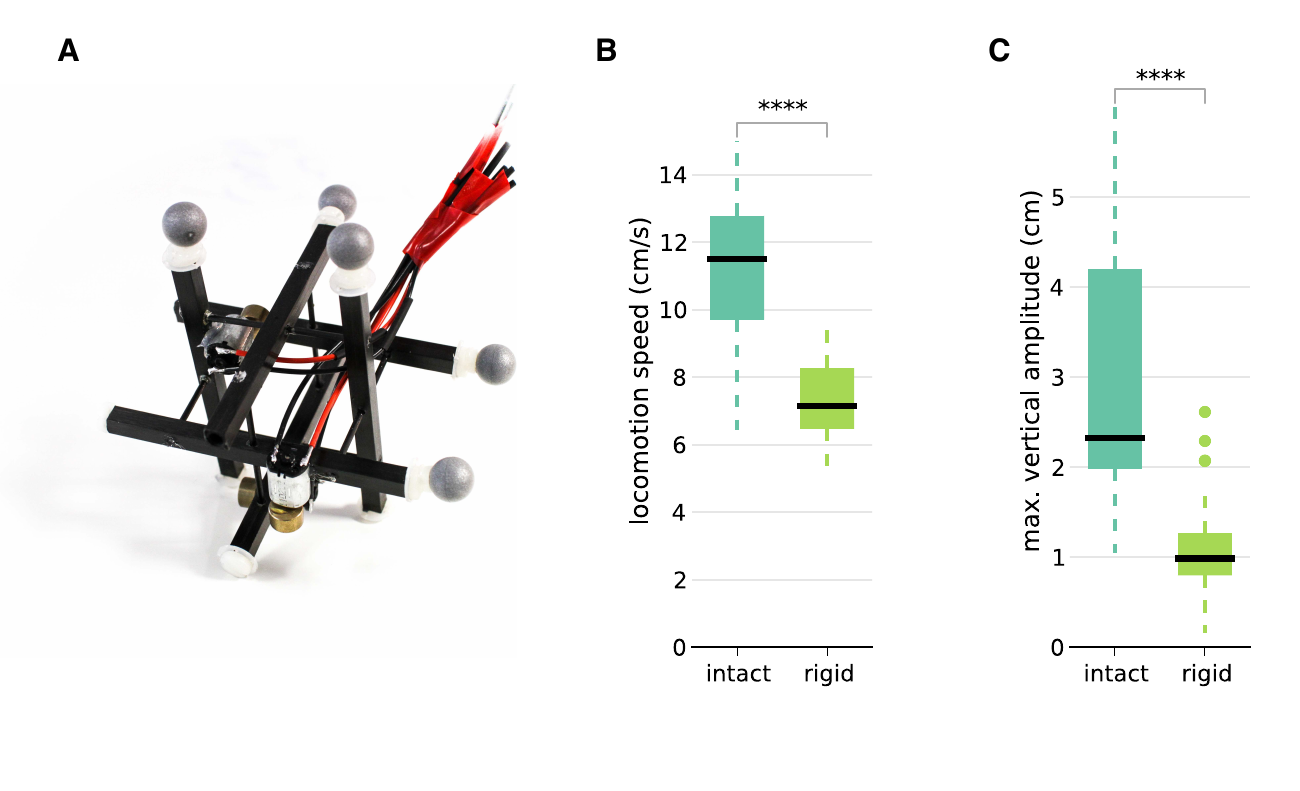}
\caption{\label{fig:rigid-res}\textbf{Experiments with a rigid robot (with priors).} \textbf{A. Rigid replica of our soft tensegrity robot.} This robot is identical to the robot of Fig.~\ref{fig:concept}, except that it contains no spring: the carbon fiber struts are held in place with small carbon fiber rods. All the dimensions, strut positions, and motor positions are the same as for the tensegrity version. \textbf{B. Locomotion speed after 30 trials for the intact (Fig.~\ref{fig:concept}) and the rigid robot (A).} Each condition is tested with 20 independent runs of the algorithm (Bayesian optimization \emph{with priors}). \textbf{C. Maximum amplitude of the markers for random gaits}. In each case, we captured the vertical position of the 4 markers for 50 random gaits of 3 seconds. We report the maximum height minus the minimum height (over the 4 markers).
For the box plots, the central mark is the median, the edges of the box are the 25\textsuperscript{th} and 75\textsuperscript{th} percentiles, the whiskers extend to the most extreme data points not considered outliers, and outliers are plotted individually. }
\end{figure*}


We then investigate our hypothesis that the interplay between a flexible tensegrity structure and vibration is the key for effective locomotion. To do so, we designed a rigid replica of our robot that does not contain any springs: the carbon fiber struts are held in place with small carbon fiber rods (Fig.~\ref{fig:rigid-res}A). All the dimensions, strut positions, and motor positions are the same as for the tensegrity version (Fig.~\ref{fig:concept}D-E). We used the same learning algorithm as for the tensegrity robot and \rev{the same prior, since we have the same intuitions about good control policies for the rigid robot as for the soft one}. We replicated the learning experiment $20$ times. The results (Fig.~\ref{fig:rigid-res}B) show that the rigid replica \rev{moves at about 60\% of the speed of the tensegrity robot}
($7.1 cm/s [5.6, 9.3]$ vs $11.5 cm/s [8.1, 13.7]$), which suggests that the flexibility of the tensegrity structure plays a critical role in its effective locomotion. In addition, we measured the amplitude of movement along the vertical axis for the end of 4 struts, both with the soft tensegrity robot and the rigid replica; we repeated this measure with 50 random gaits in both cases. These measurements (Fig.~\ref{fig:rigid-res}C) show that the markers move at least twice more when the structure is flexible ($2.3 [1.5, 4.8]$ cm vs $0.99 [0.61 , 2.1]$ cm), which demonstrates that the structure amplifies the movements induced by the vibrators.

In addition to being deformable, tensegrity structures often maintain most of their shape when a link (a spring or a strut) is missing, leading to relatively smooth failure modes. We evaluate the ability of our robot to operate after such damage by removing a spring (Fig.~\ref{fig:broken-res}A). As the shape of the robot is changed, we relaunch the learning algorithms. The experiments reveal that successful, straight gaits can be found in $30$ trials, although they are significantly lower-performing than gaits obtained with the intact robot ($11.5 cm/s [8.1, 13.7]$ versus $6.5 cm/s [5.6, 8.2]$ Fig.~\ref{fig:broken-res}B).

\begin{figure*}
  \centering
\includegraphics[width=\linewidth]{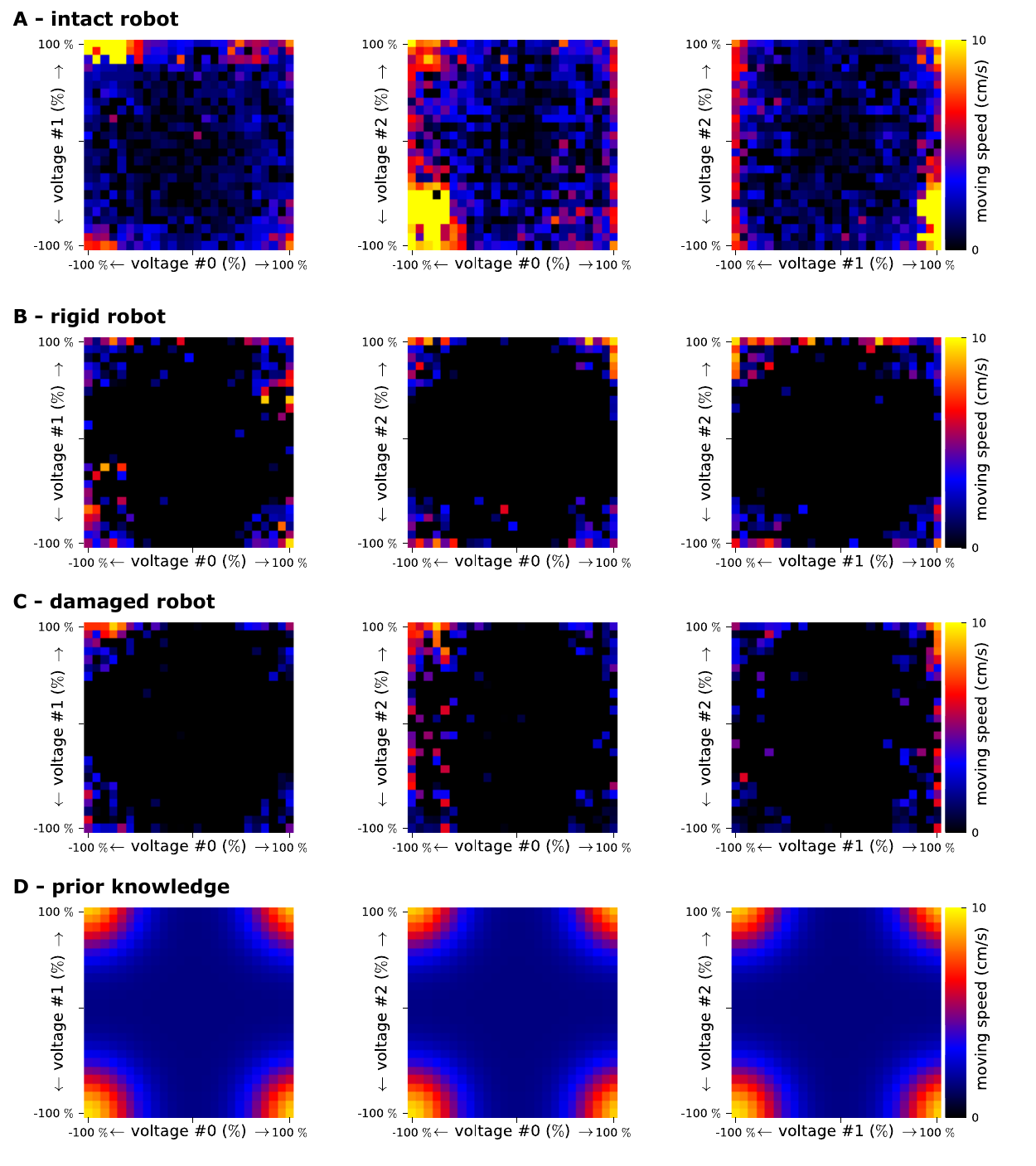}
\caption{\label{fig:profiles}\textbf{Performance profiles for all the conditions.} These performance profiles show the performance potential of each combination of 2 motor speeds (the third motor is considered as a ``free variable''). Three plots are required to get a comprehensive picture of the performance space: $v_1$ vs $v_2$, $v_1$ vs $v_3$, and $v_2$ vs $v_3$. \textbf{A. Intact robot (Fig.~\ref{fig:concept}D)}. The profiles are computed with 1800 policy evaluations (20 replicates $\times$ 30 trials  $\times$ 3 sets of experiments -- with prior, without prior, random search). \textbf{B. Rigid robot (Fig \ref{fig:rigid-res}A).} The profiles are computed with 600 policy evaluations (30 trials $\times$ 20 replicates). \textbf{C. Damaged robot (Fig.~\ref{fig:broken-res})}. The profiles are computed with 600 policy evaluations (30 trials $\times$ 20 replicates). \textbf{D. Prior knowledge.} Prior knowledge used to guide the learning algorithm (Methods).}
\end{figure*}

During all the reported experiments, we evaluated $20 \times 30 \times 3 = 1800$ different gaits on the intact robot, $20 \times 30 = 600$ gaits on the rigid robot (20 replicates, 30 trials for each replicate, and 3 treatments), and $20 \times 30 = 600$ gaits on the damaged robot. We can use these points to draw a picture of the search space that does not depend on the learning algorithm (Fig.~\ref{fig:profiles}).
Since the search space is too high-dimensional to be easily visualized (3 dimensions + performance, resulting in a 4D plot), we compute performance profiles  \citep{mouret_illuminating_2015,Reuillon2015}: for each combination of $2$ motor speeds $\big\{v_1, v_2\big\}$, we report the best performance measured regardless of the speed of the third motor (Methods).
The performance profiles (Fig.~\ref{fig:profiles}A) for the intact robot reveal that there are two high-performing regions, roughly positioned around $\big\{-100\%, 100\%,-100\%\big\}$ and $\big\{-100\%, -100\%, 100\%\big\}$ and that the first region ($\big\{-100\%, 100\%,-100\%\big\}$) is where most high-performing solutions can be found.
This finding is consistent with the prior given to the learning algorithm (Fig.~\ref{fig:profiles}D), which models that the best performance should be obtained with a combination of $-100\%$ and $+100\%$ values.
It should be emphasized that the best gaits do not correspond to the most extreme values for the motor speeds: the most reliable optima is around $\big\{-90\%, 100\%, -90\% \big\}$, mostly because too extreme values tend to make the robot tumble.
The best solutions for the rigid robots are also found in the corners, that is, for combinations of $+100\%$ and $-100\%$ motor speeds, but the measurements suggest that the optimum might be different from the one obtained with the intact robot (more data would be needed to conclude).
The data for the damaged robot show more clearly that the best solutions are around $\big\{-100\%, -100\%, 100\%\big\}$, which corresponds to the second optimum found for the intact robot (the lowest performing one).

The performance profiles thus demonstrate that the prior knowledge given to the learning algorithm is consistent with the three different robots (intact, rigid, and damaged), which suggests that it might be helpful in other situations (e.g., different damage conditions). They also demonstrate that gaits that work the best on the intact robot do not work on the damaged robot (Fig.~ \ref{fig:profiles} A versus C, second column): this shows that the learning algorithm is needed to adapt the gait if the robot is damaged.

\section*{Discussion}
Soft tensegrity robots are highly resilient, easy to assemble with the current technology, and made with inexpensive materials. \rev{In summary, vibratory soft tensegrity robots recast most of the complexity of soft robotics -- building and actuating soft structures -- into a much simpler class of robots -- easy to build and to actuate -- while keeping many of the attractive properties of soft robots -- e.g., resilience, deformability.} Thanks to the learning algorithm, our prototype can achieve locomotion speeds of more than 10 cm/s (more 1 body length per second) and learn new gaits in fewer than 30 trials, which allows it to adapt to damage or new situations. To our knowledge, this places it among the fastest soft robots.  Our soft tensegrity robots achieve this speed because they uniquely harness the flexibility and the resonance of tensegrity structures.  Discovering methods of exploiting flexibility and resonance in this manner opens new research avenues for future tensegrity structures, in particular when mechanical design can be coupled with machine learning algorithms that automatically identify how to control the resonances.


\rev{
Although our soft tensegrity robots also to large extent benefit from anisotropic friction, our effort is  distinct other vibration-based robots such as Kilobot~  \citep{rubenstein2012kilobot} and RatChair~  \citep{parshakova2016ratchair} in several important ways.  First, because of the nature of the structure, opposing pairs of vibrating motors aren't effective - and as our results show, small changes to our robot's motor speeds can have large and non-linear effects upon its behavior.  This renders the linear-interpolation approach of that work ineffective.  As a consequence, rather than relying upon hand-tuning, we instead employ Bayesian Optimization in order to determine the most effective vibrational frequencies in a minimum number of physical trials. }

\rev{
Another distinction is that our soft tensegrity robot's intrinsic resonance is tuned to respond the vibratory input of their actuators.  The benefit of this tuned resonance is particularly noticeable when the performance of the soft tensegrity robot is compared to the rigid mock-tensegrity robot described in experiments (Fig. \ref{fig:rigid-res}). These soft tensegrity robots also stand in contrast to other more rigid tensegrity robots  \citep{caluwaerts2014design,koizumi2012rolling}, which generally try to suppress their resonance. Harnessing flexibility and resonance opens new research avenues for future soft robots, in particular when mechanical design can be coupled with machine learning algorithms that automatically identify how to control the resonances.
}

\rev{
One of the more thought-provoking illustrations of the potential of soft tensegrity robots is best observed on the supplementary video, at slow speed: once properly tuned by the learning algorithm, the vibrations induce large, visible deformations of the structures that create a step-like pattern for the ``feet'' at the end of the rigid struts (more quantitative results can be seen on Fig. \ref{fig:rigid-res}-C). These step-like patterns have the potential to allow tensegrity robots to step over small irregularities of the ground like a walking robot. Importantly, these patterns are made possible by the mix of soft and rigid elements in the same structure: they are likely to be much harder to induce and control both with a fully soft robot and with a fully rigid robot. A promising research avenue is to focus on how to control the movement of the feet explicitly and make steps that are little disturbed as possible by the irregularities of the floor.
}


An added benefit of vibrational locomotion for soft robotics is that, \rev{although our current robot is tethered, it could in principle be } easy to power soft tensegrity robots with an embedded battery, by contrast with the many fluid-actuated soft robots  \citep{lipson2014challenges,shepherd2011multigait}, which need innovative ways to store energy  \citep{wehner2016integrated}. Nevertheless, soft tensegrity robots could excite their structure by other means; for instance, a flywheel that is rapidly decelerated could help the robot to achieve fast movements  \citep{romanishin2013m}, or high-amplitude, low-frequency oscillations could be generated by moving a pendulum inside the structure  \citep{chase2012review}.

\rev{Early work of ours on mobile tensegrities~  \citep{khazanov2014evolution,khazanov2013exploiting} used a rather simple interactive hill-climber in order to discover effective locomotive gaits, however this type of simplistic stochastic search was suboptimal. While there may be little qualitative difference between our earlier gaits and those described here, there are profound differences in terms of the time and data efficiency of this Bayesian Optimization approach. Most significantly, the hill-climber places no emphasis on reducing the number of physical trials performed, and as a consequence required hundreds of trials and hours of experimentation} before discovering effective gaits.  These repeated physical trials put unnecessary wear on the robot, and required a substantial amount of human effort in resetting the robot between trials.  Furthermore, the OpenCV-based optical tracking of the robot was rudimentary and lacked the spatial precision required of more effective algorithms.  The Bayesian Optimization approach we have used here, along with the high precision Optitrack system, profoundly reduces the number of physical trials and the corresponding wear on the robot, thereby increasing its capacity for faster and more autonomous resilience and adaptivity.

We purposely designed the robot so that the search space is as small as possible, which, in turn, makes it more likely for the robot to be capable of adapting in a few trials. Put differently, one of the main strength of vibration-based locomotion is to make the search problem as simple as possible.
Although, in principle, a variety of optimization techniques (e.g., simulated annealing \citep{kirkpatrick1983optimization}) might have been used, there are compelling reasons why our adaptation algorithm is based on Bayesian optimization, namely because (i) it is a principled approach to optimize an unknown cost/reward function when only a few dozen of samples are possible \citep{shahriari2016taking} (by contrast, the simulated annealing algorithm relies on the statistical properties of the search space, which are valid only with a large number of samples \citep{kirkpatrick1983optimization}), (ii) it can incorporate prior knowledge in a theoretically sound way (including trusting real samples more than prior information) \citep{cully2015robots}, and (iii) it takes into account account the acquisition noise \citep{Rasmussen2006}. For instance, Bayesian optimization is the current method of choice for optimizing the hyper-parameters of neural networks \citep{shahriari2016taking,snoek2012practical}, because evaluating the learning abilities of a neural network is both noisy and time-intensive. The downside of Bayesian optimization is a relatively high computational cost: the next sample is chosen by optimizing the acquisition function, which typically requires using a costly, non-linear optimizer like DIRECT~\citep{finkel2003direct} or CMA-ES~\citep{hansen2003reducing} (our implementation uses CMA-ES, see Methods). Put differently, Bayesian optimization trades data with computation, which makes it data-efficient, but computationally costly. As we mostly care about data-efficiency, we neglect this cost in this work, but it could be an issue on some low-power embedded computers.

Most black-box optimization (e.g. CMA-ES \citep{hansen2003reducing}) and direct policy search algorithms (e.g policy gradients   \citep{peters2006policy}), could substitute Bayesian optimization as an adaptation algorithm by directly optimizing the reward (instead of first modeling it with Gaussian process). While they would not need time-intensive optimizations to select the next sample to acquire, these algorithms are tailored for at least a thousand evaluation (e.g. $10^4$ to $10^5$ evaluations in benchmarks of 2D functions for black-box optimizers   \citep{hansen2010comparing}), are not designed to incorporate priors on the reward function, and are, at best, only tolerant to noisy functions. As a consequence, while algorithms like CMA-ES could work as an adaptation algorithm, they appear to be a sub-optimal choice for online adaptation when only a few dozen of evaluations are possible.

Traditional Bayesian optimization uses a constant mean as a prior \citep{lizotte2007automatic,calandra2014experimental}, that is, the only prior knowledge is the expectation of the cost/reward. By contrast, we show here that it is effective to introduce some basic intuitions about the system as a non-constant prior on the reward function. We thus increase the data-efficiency while keeping the learning algorithm theoretically consistent. Cully \emph{et al.}  \citep{cully2015robots} also used a non-constant prior; however, (i) they generated it using a physics simulator, which is especially challenging for a vibrating tensegrity robot, and (ii) they only computed this prior for a discrete set of potential solutions, which, in turn, constrain Bayesian optimization to search only in this set. Here we follow a more continuous approach as our prior is a continuous function, and we show that relevant priors can be defined without needing a physics simulator. The more general problem of how to generate ``ideal'' priors is far from trivial. Intuitively, priors should come from a meta-learning process  \citep{lemke2015metalearning}, for instance, an evolution-like process \citep{kirschner2006plausibility}, which would search for priors that would work well in as many situations as possible (i.e., instincts). Effectively implementing such a process remains an open grand challenge in machine learning and is sometimes called ``meta-learning'' \citep{lemke2015metalearning}.

Putting all these attractive features altogether, soft tensegrity robots combine simplicity, flexibility, performance, and resiliency, which makes this new class of robots a promising building block for future soft robots. 

Of course, additional work is needed to have a more complete theory of the ``optimal suppleness'' of soft robots. Intuitively, too much suppleness would absorb the energy transmitted by the vibrator and prevent effective gaits; but, at the other end of the spectrum, a rigid robot cannot generate the form changes that are necessary for the most interesting gaits. This may be what we observed when we damaged the robot: by making the structure less constrained, the shape of the robot may have become looser, and ``softer'', which impacted the maximum locomotion speed (alternatively, the removal of a spring might have prevented the transmission of some oscillations or some resonance modes).  Nevertheless, for every kind of suppleness that we tried, the Bayesian optimization algorithm was always capable of finding some effective gaits, which means that the ``optimal softness'' does not need to be known {\em a priori} in order to discover effective locomotion. In this regard, trial-and-error approaches like the ones used here provide a valuable ability to respond and adapt to changes online in a rather robust manner, much like living systems~\citep{cully2015robots}.

Several exciting open questions remain.  So far, we have only demonstrated the effectiveness of this technique on a single substrate rather than across an entire range of environments. A compelling question we look forward to exploring in future work, for instance, is the extent to which the locomotive gaits we have discovered are robust and self-stabilizing in the face of external perturbations and changes in the substrate.  Of course, the general problem of robust locomotion of any robot, much less soft robots, across multiple substrates and environments remains a relatively open topic.  Recent work has, for instance, explored hand-picked strategies for the quasi-static locomotion of a cable-actuated tensegrity on inclined surfaces~\citep{incline}.   Our own ability to harness tensegrity vibration in order to induce large-scale and dynamic structure offers a compelling and promising method of discovering much more dynamic gaits for these environments. Indeed, our robot design is already capable of interesting behavioral diversity, including several unique rolling behaviors, which might be beneficial across environments - however we were unable to explore these more deeply due to the tethered nature of this design.  Nonetheless, the speed with which our algorithm can learn effective gaits, especially when damaged, provides a glimpse into how future soft robots could adapt to new and unexpected environments {\em in situ}, with no pre-existing knowledge or experience of that environment.

\begin{figure}
  \centering
\includegraphics[width=\linewidth]{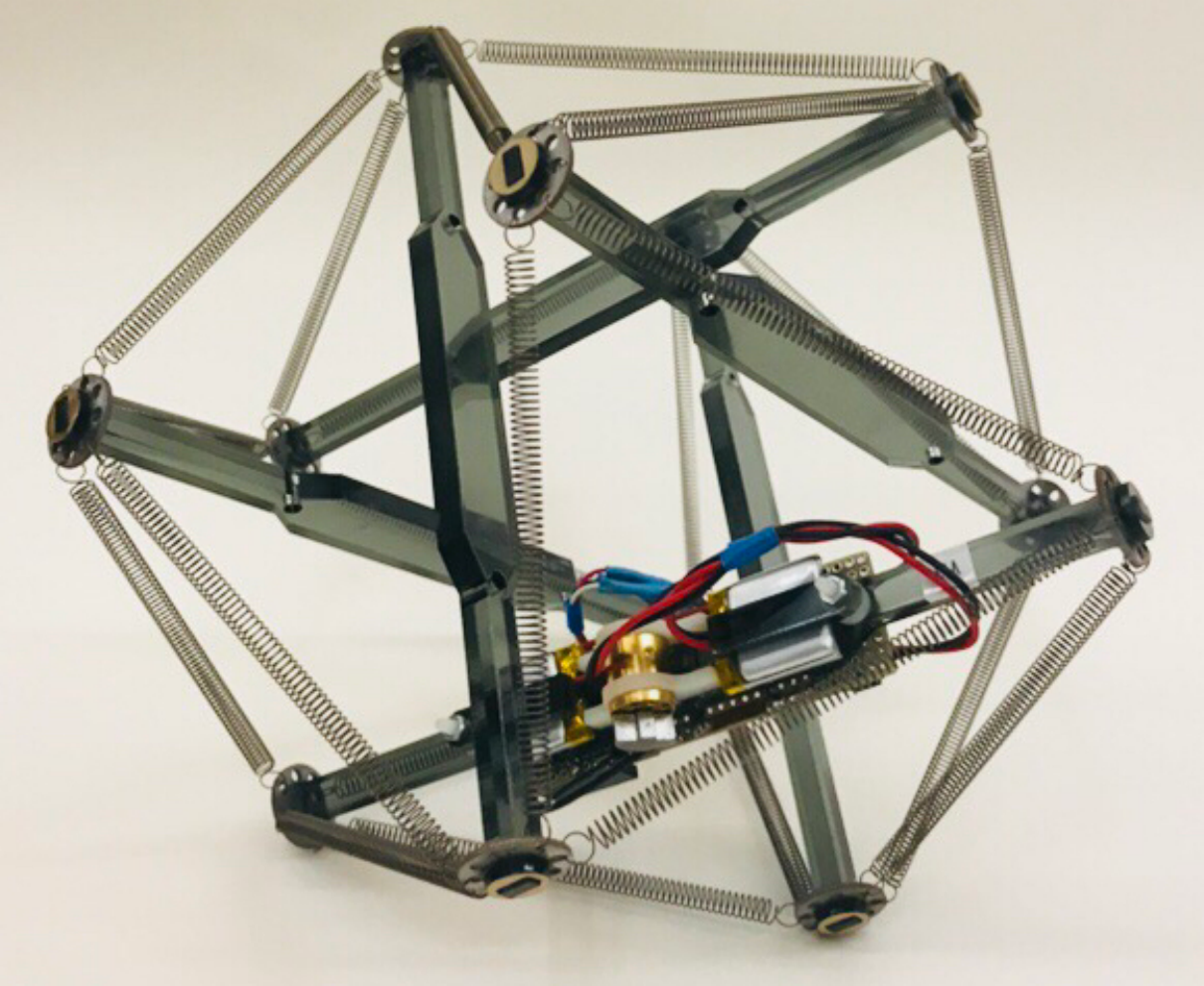}
\caption{\label{fig:wireless}\textbf{An untethered version of the tensegrity robot.}  This new robot, still under development will allow for more interesting dynamical behaviors such as rolling, as well as complex environments.  This could in principle allow for completely on-board learning as well.}
\end{figure}

This leads to the recognition that the the present prototype, while more than sufficient to demonstrate the claims of this paper, is not yet fully autonomous : it relies on a tether for power, uses an external motion capture to evaluate its performance (locomotion speed), and uses an offboard computer for the the learning algorithm.
We are in the process of designing a fully wireless and autonomous tensegrity robot, as illustrated by Figure~\ref{fig:wireless}.  This next generation of robot will be capable of substantially more dynamical behaviors, such as rolling and jumping, and more capable of exploring complex environments.  Evaluating the performance of locomotion techniques using on-board processing could in principle by achieved either with accelerometers or with an embedded camera paired with a visual odometry algorithm, \citep{forster2014svo,cully2015robots}, but the vibrations and the fast movements of the struts are likely to disturb many visual algorithms.  Additionally, the modular nature of this wireless strut design means that we could explore an entire range of tensegrity robot morphologies, including those with considerably more than six struts.

Overall, our soft tensegrity robots move thanks to the complex interactions between the actuators (vibrators), the structure (springs and struts), and the environment (the ground). This kind of emergent behavior is central in the embodied intelligence theory  \citep{pfeifer2007self}, which suggests that we will achieve better and more life-like robots if we encourage such deep couplings between the body and the ``mind'' -- here, the controller. However, as demonstrated in the present work, trial-and-error learning algorithms offer a strongly viable approach to discovering these emergent behaviors.

~\\ 

\section*{Material and Methods}

\subsection*{Robot} The tensegrity used is defined by six equal length composite struts which are connected to each other via 24 identical helical springs, with four springs emanating from each strut end.  This follows the geometry described as TR-6 by Skelton \citep{skelton2009tensegrity}.  Few actual machining operations are required to produce the  tensegrity. The six 9.4 cm long composite struts are cut from 6.35 mm square graphite composite tubes (Goodwinds).  The three 12mm vibrational motors (Precision Microdrives Model 312-107)  were mounted to the flat outer surface of the struts using hot melt adhesive.  Both ends of each strut were then tapped for 10-24 nylon screws fitted with nylon washers. The hooked ends of the helical springs (Century Spring Stock No. 5368) were attached directly to holes drilled through the nylon washers.  The motors were connected via thin gauge magnet wire to Serial Motor Controllers (Pololu Qik 2s9v1 Dual Serial Motor Controller) connected in turn to a USB Serial Adapter (SparkFun FTDI Basic Breakout board)

\rev{The specific spring constants were chosen in order to produce relatively low natural frequencies and correspondingly large displacements of the structure while at the same time limiting estimated static deflection to 5\% of strut length.    In order to determine this, a single strut was modeled as being connected to four linear springs at each end, equally spaced around the radius, each at a $45^{\circ}$ angle.  Limiting static deflection to 5\% of strut length results in a spring constant value of 0.209 N/cm. Subsequently, the entire 6-bar structure was modeled by assuming that one strut was be anchored in place and then using matrix structural analysis to determine the natural frequencies.  The vibrational motor was then chosen that was capable of generating sufficient centrifugal force at a suitable range of frequencies.  Details of the modeling and design are provided in ~  \citep{khazanov2013exploiting}.}

\subsection*{Control policy} Each policy is defined by three PWM values that determine the input voltage of the 3 vibrating motors ($\mathbf{\chi} = [v_1, v_2, v_3]$), which can take values between $0$ (full speed, backward) and $1$ (full speed, forward); $0.5$ corresponds to a speed of $0$, that is, to no movement.

\subsection*{Performance function} Each controller is tested for $3$ seconds, then the Euclidean distance between the starting point and the end point is recorded. The performance function is the distance (in cm/s) divided by 3. If during the $3$ second evaluation period the yaw of the robot exceeds 1 radian, the evaluation is stopped and the recorded distance is the distance between the starting point and the point reached by the robot when it exceeded the yaw limit.

The policies are evaluated externally with a motion tracking system (Optitrack Prime 13 / 8 cameras), but the same measurements can be obtained with an embedded camera connected to a visual odometry system  \citep{davison2007monoslam,cully2015robots}.

\subsection*{Profile plots} We use the profile plots to depict the search space and the prior used by the learning algorithm (Fig.~\ref{fig:profiles}). For each pair of dimensions, we discretize the motor speeds into $25$ bins. For each bin, we compute $p_{profile}(v_1, v_2) = \max_{v_3} p(v_1, v_2, v_3)$, where $p(v_1, v_2, v_3)$ is the performance of the robot for motor speeds $v_1, v_2, v_3$ and $p_{profile}(v_1, v_2)$ is the performance reported in the profile. To get a comprehensive pictures, we need three plots: $p_{profile}(v_1, v_2)$, $p_{profile}(v_1, v_3)$, and $p_{profile}(v_2, v_3)$.

\subsection*{Learning algorithm}
Our learning algorithm allows the robot to discover by trial-and-error the best rotation speeds for its three motors. It essentially implements a variant of Bayesian optimization, which is a state-of-the-art optimization algorithm designed to maximize expensive performance functions (a.k.a. cost functions) whose gradient cannot be evaluated analytically  \citep{ghahramani2015probabilistic,shahriari2016taking}. Like other model-based optimization algorithms (e.g., surrogate-based algorithms  \citep{booker1999rigorous,forrester2009recent,jin2011surrogate}, kriging  \citep{simpson1998comparison}, or DACE  \citep{jones1998efficient,sacks1989design}), Bayesian optimization models the objective function with a regression method, uses this model to select the next point to acquire, then updates the model, etc. until the algorithm has exhausted its budget of function evaluations.

Here a Gaussian process models the objective function  \citep{Rasmussen2006}, which is a common choice for Bayesian optimization  \citep{calandra2014experimental, brochu2010tutorial, lizotte2007automatic,ghahramani2015probabilistic,shahriari2016taking}. For an unknown cost function $f$, a Gaussian process defines the probability distribution of the possible values $f(\mathbf{x})$ for each point $\mathbf{x}$. These probability distributions are Gaussian, and are therefore defined by a mean ($\mu$) and a variance ($\sigma^2$). However, $\mu$ and $\sigma^2$ can be different for each $\mathbf{x}$; a Gaussian process therefore defines a probability distribution \emph{over functions}:
 \begin{equation}
P(f(\mathbf{x})|\mathbf{x}) = \mathcal{N}(\mu(\mathbf{x}), \sigma^2(\mathbf{x}))
\end{equation}
where $\mathcal{N}$ denotes the standard normal distribution.

At iteration $t$, if the performance $[P_{1},\cdots,P_{t}]=\mathbf{P}_{1:t}$ of the points $[\mathbf{\chi}_1, \cdots, \mathbf{\chi_t}]=\mathbf{\chi}_{1:t}$ has already been evaluated, then $\mu_t(\mathbf{x})$ and $\sigma_t^2(\mathbf{x})$ are fitted as follows  \citep{Rasmussen2006}:

\begin{equation}\label{eq:GP}
\begin{gathered}
\begin{array}{l}
 \mu_{t}(\mathbf{x})= \mathbf{k}^\intercal\mathbf{K}^{-1}\mathbf{P}_{1:t}\\
 \sigma_{t}^2(\mathbf{x})=k(\mathbf{x},\mathbf{x}) + \sigma_{noise}^2 - \mathbf{k}^\intercal\mathbf{K}^{-1}\mathbf{k}\\
 \textrm{where:}\\
 \mathbf{K}=\left[ \begin{array}{ c c c}
    k(\mathbf{\chi}_1,\mathbf{\chi}_1) &\cdots & k(\mathbf{\chi}_1,\mathbf{\chi}_{t}) \\
    \vdots   &  \ddots &  \vdots  \\
    k(\mathbf{\chi}_{t},\mathbf{\chi}_1) &  \cdots &  k(\mathbf{\chi}_{t},\mathbf{\chi}_{t})\end{array} \right]
+ \sigma_{noise}^2I\\
 \mathbf{k}=\left[ \begin{array}{ c c c c }k(\mathbf{x},\mathbf{\chi}_1) & k(\mathbf{x},\mathbf{\chi}_2) & \cdots & k(\mathbf{x},\mathbf{\chi}_{t}) \end{array} \right]
 \end{array}
\end{gathered}
\end{equation}

The matrix $\mathbf{K}$ is called the covariance matrix. It is based on a \emph{kernel function} $k(\mathbf{x_1}, \mathbf{x_2})$ which defines how samples influence each other. Kernel functions are classically variants of the Euclidean distance. Here we use the \emph{exponential kernel}  \citep{Rasmussen2006,brochu2010tutorial,shahriari2016taking,cully2015robots}:
\begin{equation}
  k(\mathbf{x_1}, \mathbf{x_2})  = \exp \Big(-\frac{1}{\beta^2} ||\mathbf{x_1} - \mathbf{x_2}||^2\Big)
  \label{eq:kernel}
\end{equation}
because this is the most common kernel in Bayesian optimization and we did not see any reason to choose a different one \citep{brochu2010tutorial,shahriari2016taking}. We fixed $\beta$ to $0.15$.

An interesting feature of Gaussian processes is that they can easily incorporate a prior $\mu_p(\mathbf{x})$ for the mean function, which helps to guide the optimization process to zones that are known to be promising:

\begin{equation}
\mu_{t}(\mathbf{x})= \mu_p(\mathbf{x}) + \mathbf{k}^\intercal\mathbf{K}^{-1}(\mathbf{P}_{1:t}-\mu_p(\mathbf{\chi}_{1:t}))
\label{eq:prior}
\end{equation}

In our implementation, the prior is a second Gaussian process defined by hand-picked points (see the ``prior'' section below).

To select the next $\mathbf{\chi}$ to test ($\mathbf{\chi}_{t+1}$), Bayesian optimization maximizes an \emph{acquisition function}, a function that reflects the need to balance exploration -- improving the model in the less known parts of the search space -- and exploitation -- favoring parts that the model predicts as promising. Numerous acquisition functions have been proposed (e.g., probability
of improvement, the expected improvement, or the Upper Confidence
Bound (UCB)  \citep{brochu2010tutorial, calandra2014experimental,shahriari2016taking}); we
chose UCB because it provided the best results in several previous
studies  \citep{brochu2010tutorial, calandra2014experimental} and because of its simplicity. The
equation for UCB is:
\begin{equation}
\mathbf{\chi}_{t+1}= \operatorname*{arg\,max}_\mathbf{x} (\mu_{t}(\mathbf{x})+ \kappa\sigma_t(\mathbf{x}))
\label{ucb}
\end{equation}
where $\kappa$ is a user-defined parameter that tunes the tradeoff between exploration and exploitation. We chose $\kappa=$ 0.2.

\subsection*{Prior for the learning algorithm} The learning algorithm is guided by a prior that captures the idea that the highest-performing gaits are likely to be a combination of motors at full speed (in forward or in reverse). In our implementation, it is implemented with a Gaussian process defined by 9 hand-picked points and whose variance is ignored (equation \ref{eq:GP}). The kernel function is the exponential kernel (equation \ref{eq:kernel}), with $\beta=0.15$.

The 9 hand-picked points ($\mathbf{\chi_1}, \cdots, \mathbf{\chi_9}$) are as follows (Fig~\ref{fig:profiles}-D):
\begin{displaymath}
  \begin{array}{l}
  \mathbf{\chi}_1 = [-100\%, -100\%, -100\%]\\
\mathbf{\chi}_2 = [-100\%, -100\%, +100\%]\\
\mathbf{\chi}_3 = [-100\%, 100\%, -100\%]\\
\mathbf{\chi}_4 = [-100\%, +100\%, +100\%]\\
\mathbf{\chi}_5 = [+100\%, -100\%, -100\%]\\
\mathbf{\chi}_6 = [+100\%, +100\%, +100\%]\\
\mathbf{\chi}_7 = [+100\%, -100\%, -100\%]\\
\mathbf{\chi}_8 = [+100\%, -100\%, +100\%]\\
\mathbf{\chi}_9 = [0\%, 0\%, 0\%];\\
P(\mathbf{\chi}_1), \cdots, P(\mathbf{\chi}_8) = 0.3;
P(\mathbf{\chi}_9) = 0
\end{array}
\end{displaymath}

\subsection*{Statistics} For all experiments, we report the $5^{th}$ and $95^{th}$ percentiles. We used a two-tailed Mann-Whitney U test for all statistical tests. \rev{For the box plots, the central mark is the median, the edges of the box are the 25\textsuperscript{th} and 75\textsuperscript{th} percentiles (inter-quartile range -- IQR), the whiskers corresponds to the range $[25\% - 1.5 \times IQR, 75\% + 1.5\times IQR]$, and points outside of the whiskers are considered to be outliers (this corresponds to the ``interquartile rule'').} For each box plot, the result of the Mann-Whitney U test (two-tailed) is indicated with stars: * means $p \leq 0.05$, ** means  $p \leq 0.01$, ***  means $p \leq 0.001$, and **** means  $p \leq 0.0001$.

\subsection*{Computer code} \url{http://members.loria.fr/JBMouret/src/limbo-tensegrity.tar.gz} ; this code will be released with an open-source license on Github for the final publication.

\subsection*{Data availability} \url{http://members.loria.fr/JBMouret/data/tensegrity.tar.gz} ; these data will be released on Dryad for the final publication.

\sffamily
\bibliography{tensegrity}

\section*{Acknowledgments}
This project received funding from the European Research Council (ERC) under the European Union's Horizon 2020 research and innovation programme (Project: ResiBots, grant agreement No 637972); J.R. received funding for a 1-month visit at University of Lorraine, France. The authors would like to thank Dorian Goepp for his invaluable help the ResiBots team for their comments on this manuscript.  The authors would also like to thank Bill Keat for his help and insight into the design of the robot, and all the undergraduates of Union College's Evolutionary Robotics Group.

The pictures of Fig.\ref{fig:concept} are distributed under a Creative Commons License. A: \textcopyright Jonathan Rieke. B: \textcopyright Margaret Donald. C: \textcopyright National Institute of Dental and Craniofacial Research, National Institutes of Health (NIH).

\section*{Author Contributions}
J.R. designed the robot; J.R. and J.-B.M. designed the study, performed the experiments, analyzed the data, and wrote the paper.

\section*{Supplementary information}
\subsection*{Video S1}
\textbf{Presentation of our soft tensegrity robot.} The video shows the soft tensegrity robot in action: how it can locomote and how it can learn to compensate when damaged.
The video is available online one: \url{https://youtu.be/SuLQDhrk9tQ}

\end{document}